\documentclass[10pt,twocolumn,letterpaper]{article}
\usepackage[table,dvipsnames]{xcolor}
\usepackage{cvpr}              

\usepackage[normalem]{ulem}
\usepackage{threeparttable}
\usepackage{multirow}
\usepackage{booktabs}
\usepackage{makecell}
\usepackage{array}
\usepackage{amsmath}

\definecolor{cvprblue}{rgb}{0.21,0.49,0.74}
\usepackage[pagebackref,breaklinks,colorlinks,allcolors=cvprblue]{hyperref}

\def\paperID{} 
\def\confName{CVPR}
\def\confYear{2025}

\title{From Pixels to Gigapixels: Bridging Local Inductive Bias and Long-Range Dependencies with Pixel-Mamba}

\author{Zhongwei Qiu$^{1,2,5,}$\thanks{Equal Contribution} , Hanqing Chao$^{1,2,6,*}$, Tiancheng Lin$^{1,2,6,*}$, Wanxing Chang$^{1,2}$,\\
Zijiang Yang$^{1,7}$, Wenpei Jiao$^{1,8}$,Yixuan Shen$^3$, Yunshuo Zhang$^4$, Yelin Yang$^4$,\\
Wenbin Liu$^3$, Hui Jiang$^4$, Yun Bian$^3$, Ke Yan$^{1,2}$, Dakai Jin$^1$, Le Lu$^1$\\
{\small$^1$DAMO Academy, Alibaba Group, $^2$Hupan Lab}\\
{\small$^3$Department of Radiology, Changhai Hospital, $^4$Department of Pathology, Changhai Hospital}\\
{\small$^5$Zhejiang University, $^6$Fudan University, $^7$University of Science and Technology Beijing, $^8$ Peking University}\\
{\tt\small \{qiuzhongwei.qzw, hanqing.chq\}@alibaba-inc.com}
}

\begin{document}
\maketitle
\begin{abstract}

Histopathology plays a critical role in medical diagnostics, with whole slide images (WSIs) offering valuable insights that directly influence clinical decision-making. However, the large size and complexity of WSIs may pose significant challenges for deep learning models, in both computational efficiency and effective representation learning. In this work, we introduce Pixel-Mamba, a novel deep learning architecture designed to efficiently handle gigapixel WSIs. Pixel-Mamba leverages the Mamba module, a state-space model (SSM) with linear memory complexity, and incorporates local inductive biases through progressively expanding tokens, akin to convolutional neural networks. This enables Pixel-Mamba to hierarchically combine both local and global information while efficiently addressing computational challenges. Remarkably, Pixel-Mamba achieves or even surpasses the quantitative performance of state-of-the-art (SOTA) foundation models that were pretrained on millions of WSIs or WSI-text pairs, in a range of tumor staging and survival analysis tasks, {\bf even without requiring any pathology-specific pretraining}. Extensive experiments demonstrate the efficacy of Pixel-Mamba as a powerful and efficient framework for end-to-end WSI analysis.


\end{abstract}    
\section{Introduction}
\label{sec:intro}

\begin{figure}[t]
  \centering
  \includegraphics[width=\linewidth,clip=true, trim=0 275 190 0]{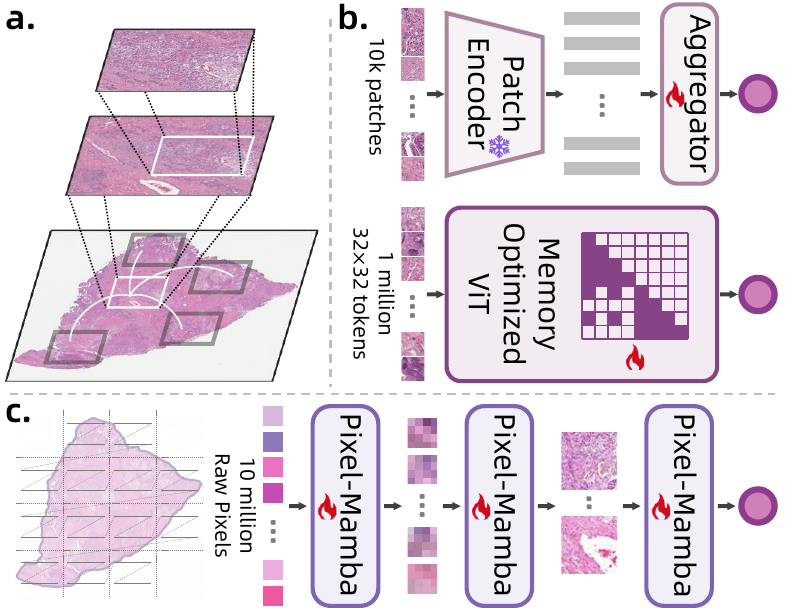}
  \vspace{-0.5cm}
  \caption{(a) Pathologists integrate observations from multiple regions across different scales to make a comprehensive assessment. (b) Frameworks of mainstream WSI analysis methods: a two-stage pipeline (top)
  and memory-optimized ViT (bottom, often with heavily pruned attention). (c) The proposed Pixel-Mamba, an end-to-end framework that combines progressive token expansion and the Mamba module to effectively integrate local inductive biases with long-range dependencies in a hierarchical manner.}
  \label{fig:fig1}
\end{figure}

Histopathology plays a pivotal role in medical diagnostics, serving as the gold standard for diagnosing and characterizing a wide range of diseases~\cite{corredor2019spatial,saltz2018spatial}. By examining tissue microstructures, pathologists obtain critical insights into the nature and severity of conditions, which directly influence treatment decisions and patient management~\cite{saltz2018spatial,corredor2019spatial}. With the advent of WSI, computational pathology, driven by artificial intelligence (AI), has shown promise in addressing many challenges in Pathology, such as survival prediction, cancer subtyping, and predicting molecular alterations~\cite{jaume2024hest}. However, the large size of WSIs, which can reach up to $50,000 \times 50,000$ pixels, presents significant computational and methodological challenges for AI models.

From a computational perspective, even at low magnification, a single WSI can contain tens of millions of pixels, making training of deep learning models \mbox{impractical}. From a representation learning standpoint, WSIs introduce two key challenges: 1) balancing local information with long-range dependencies and 2) managing hierarchical structures across multiple scales. 
For example, \mbox{tertiary} lymphoid structures (TLS), which are strongly associated with favorable prognosis in various cancers, span only $\sim50,000 \mu m^2$ (roughly $\sim2,000$ pixels at 2× magnification). However, analyzing their distribution requires considerably larger regions, approximately 1 $cm^2$ or up to $4$ million pixels at the same magnification, due to tumor heterogeneity and the sparse nature of TLS~\cite{barmpoutis2021tertiary,van2024multi}. Moreover, WSIs encapsulate hierarchical information, with multi-scale contexts intricately intertwined. For tasks like diagnosing gastric signet ring cell carcinoma, both low- and high-magnification information are indispensable: large-scale regions at low magnification help detect potential tumor areas, while high-magnification is essential for examining cell morphology. Without effective integration of multi-scale information, errors such as misclassifying signet ring cells as foam or mucinous cells could lead to severe diagnostic errors~\footnote{Using low magnification alone may lead to confusion with foam cell clusters in certain cases, whereas relying solely on high magnification may result in confusion with mucinous cells.}.


Most existing WSI analysis methods follow a two-stage pipeline~\cite{chen2022scaling,xu2024whole}. First, WSIs are divided into small patches (e.g., $256\times 256$ pixels), and a pre-trained encoder is used to extract features from patches. Second, these features are aggregated using methods such as multiple instance learning (MIL) to predict WSI-level outcomes. While this approach addresses the computational challenges of WSIs, it disrupts the flow of multi-scale information by isolating local and global features into separate stages. Consequently, hierarchical relationships across scales are lost during training, limiting the model’s capacity. A few recent efforts, such as LongViT~\cite{wang2023image}, aim to maintain end-to-end training by optimizing Transformer memory usage. However, the massive number of tokens in gigapixel images, combined with their heavily pruned attention, hinder their performance.


To address these challenges, we propose Pixel-Mamba, a novel architecture designed for efficient and comprehensive WSI analysis. To tackle the computational burden, we leverage the Mamba module, a type of SSM with linear memory complexity, which has demonstrated exceptional performance in modeling ultra-long sequences~\cite{gu2023mamba,zhu2024vision}. 
Furthermore, we argue that local inductive biases are particularly critical for WSI analysis. Small and spatially adjacent structures (e.g., cells) often combine to form meaningful larger structures (e.g., blood vessels), and similar local patterns can guide long-range dependencies. 
For instance, while TLS structures may be spatially distant, they are often morphologically similar. Pixel-Mamba integrates these insights by combining progressive token expansion with the Mamba module. 
Starting from pixel-level tokens, our network progressively expands the token receptive field across layers, reaching regions as large as $32\times32$. This design introduces local inductive biases, enabling the effective learning of hierarchical information, akin to convolutional neural networks, while the Mamba module further ensures that long-range dependencies are modeled at every level. It allows the network to integrate both local and global information hierarchically, addressing both computational and methodological challenges. 
Our experiments demonstrate the effectiveness of Pixel-Mamba across a variety of datasets and tasks. 
Notably, without requiring any pathology-specific pre-training, Pixel-Mamba matches or even exceeding the performance of leading foundation models (FM) pre-trained on millions of WSIs or WSI-text pairs. This underscores the potential of Pixel-Mamba as a powerful and efficient framework and a simple practical baseline for end-to-end WSI analysis.
Our main contributions can be summarized as follows:
\begin{itemize}
    \item We propose Pixel-Mamba, a high-memory efficiency architecture designed for end-to-end WSI analysis.
    \item Pixel-Mamba introduces progressive tokens expansion to incorporate hierarchical local inductive biases, concurrently models long-range dependencies across all scales, and enables efficient modeling of gigapixel WSIs.
    \item Extensive experiments demonstrate that Pixel-Mamba achieves or surpasses the performance of SOTA FMs pre-trained on millions of WSIs or WSI-text pairs, without requiring any pathology-specific pre-training.
\end{itemize}

\section{Related Work}
\label{sec:related_work}

\subsection{Representation Learning in WSIs}

\paragraph{Two-stage MIL:}
Existing MIL methods adopt a two-stage scheme. First, they divide WSIs into patches of fixed size using pre-processing tools like CLAM \cite{lu2021data}, and extract patch features using pre-trained feature extractors such as ResNet \cite{he2016deep}, ViT \cite{alexey2020image}, or pre-trained foundation models \cite{wang2021transpath,xu2024whole,chen2024uni,lu2024visual}.
Most two-stage MIL methods focus on the second stage and aim to explore better strategies for aggregating patch-level representations into slide-level representations. 
With the success of Transformers~\cite{qiu2022ivt,qiu2023learning}, the attention mechanism has been introduced to aggregate patch features. Attention-based MIL approaches aim to capture long-sequence relationships among tens of thousands patches through attention mechanisms, including ABMIL \cite{ilse2018attention}, TransMIL \cite{shao2021transmil}, ILRA-MIL \cite{xiang2023exploring}, and others~\cite{yao2020whole,li2021dual,zhang2022dtfd}.
Due to the difficulties in capturing long-range dependencies, state space models have been used to model the relationships among patches by \cite{fillioux2023structured, yang2024mambamil}.

\paragraph{Hierarchical Representation:}
Current methods usually only use the WSIs at 20$\times$ magnification, while some works take advantage of the multi-scale nature of WSIs. 
They can be roughly divided into two categories, depending on how to process the multi-scale information. Some works are parallel, for example, multi-resolution input~\cite{multi-input1,multi-input2}, multi-scale feature concatenation~\cite{li2021dual}, and cross-resolution graph construction~\cite{h2mil} are proposed to explore multi-scale representations simultaneously.
Other works are cascaded: HIPT~\cite{HIPT} and GigaPath~\cite{gigapath} use different ViTs to encode tile-level and slide-level features stage by stage, where the output of the first stage is fed into the next stage. Similar works include those~\cite{pathologist1,pathologist2}.

\paragraph{End-to-end Methods:}
To optimize patch embedding for downstream tasks, ICMIL~\cite{wang2024rethinking} attempts to fine-tune patch feature extractors in the MIL framework by establishing loss propagation from the classifier to the feature extractor through iterative training. However, the feature extractor and task head is not synchronously optimized. VPTSurv~\cite{qiu2024end} uses adaptors to transfer pre-trained models to WSI. LongViT~\cite{wang2023image} builds an end-to-end framework using LongNet~\cite{ding2023longnet}, a vision Transformer with dilated attention that is pre-trained using 64 GPUs. Despite this, LongViT's performance still lags behind two-stage MIL approaches with pre-trained foundation models and fails to capture hierarchical representations for WSIs.
In this work, we aim to develop a lightweight and efficient end-to-end framework for WSI analysis.

\subsection{Pathological Foundation Models}
Most existing approaches for pathological analysis are based on the two-stage MIL framework, making the design and pre-training of the feature extractor crucial. Early methods~\cite{lu2021data,ilse2018attention} employed the ResNet~\cite{he2016deep} or Swin Transformer~\cite{liu2021swin} pre-trained on natural images like ImageNet as the feature extractor. To adapt these extractors to pathological images, other works have designed multiple pretext tasks~\cite{koohbanani2021self,yang2021self}, contrastive learning strategies~\cite{li2021dual,huang2021integration}, and generative modeling methods~\cite{luo2023self,quan2024global} for pre-training. Recently, with the expansion of data scale and the development of multimodal learning, new pre-trained models, often referred to as foundation models (FMs), have been released. Typical FMs include CTransPath~\cite{wang2021transpath}, PLIP~\cite{huang2023visual}, UNI~\cite{chen2024uni}, CONCH~\cite{lu2024visual}, GigaPath~\cite{xu2024whole}, and CHIEF~\cite{wang2024pathology}. Although these FMs can extract better patch embeddings for MIL, the limitations of the MIL framework and the fixed feature embeddings still constrain their performance.

\subsection{Mamba in Computational Pathology}
Compared with Transformer, state space models such as Mamba~\cite{gu2023mamba,dao2024transformers} exhibit linear computational complexity for long sequence modeling and have been used as a new vision backbone, exemplified by~\cite{zhu2024vision,liu2024vmamba}. Recently, Mamba has been introduced into computational pathology to model the long sequence relationships among image patches~\cite{fillioux2023structured,yang2024mambamil} or as a patch extractor~\cite{nasiri2024vim4path}. Although they have achieved certain performance improvements, they are still constrained by the limitations of the MIL framework. This paper leverages the linear computational complexity of Mamba and proposes a novel end-to-end framework for learning slide-level representations of pathological images.


\section{Method}
\label{sec:method}

\begin{figure*}[t]
  \centering
  \includegraphics[width=\textwidth,clip=true, trim=0 15 5 0]{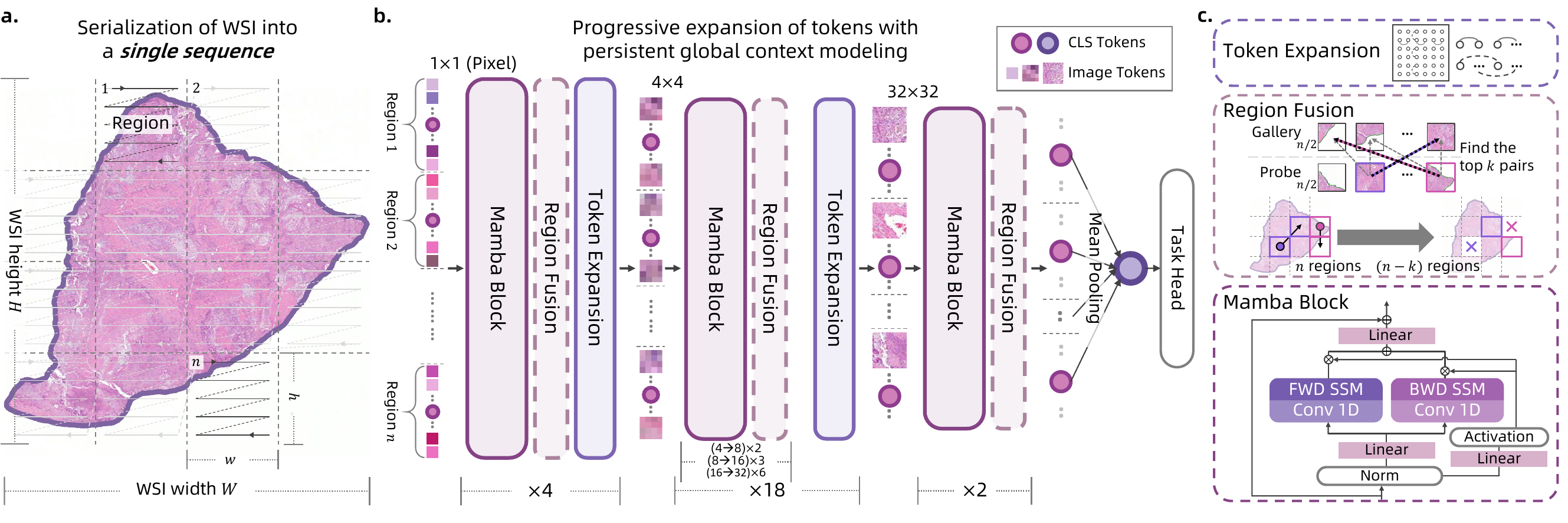}
\vspace{-0.5cm}
  \caption{The Pixel-Mamba Framework. (a) The WSI is serialized, with CLS tokens added to create the token series $T$. (b) Pixel-Mamba progressively expands the receptive field of tokens while maintaining global context modeling. (c) Detailed illustrations of the Mamba Block, Region Fusion, and Token Expansion modules.
  }
  \label{fig:architecture}
\end{figure*}

\subsection{Framework}
The Pixel-Mamba framework, illustrated in Figure ~\ref{fig:architecture}, consists of three main steps: whole slide image serialization, the Pixel-Mamba network, and downstream task heads.
To address the challenges associated with gigapixel WSIs, Pixel-Mamba begins by serializing each pixel of the WSI into tokens through a whole slide scan. The Pixel-Mamba network then processes these tokens layer by layer, progressively expanding their receptive fields to capture hierarchical information. At each level, it employs a state space model (SSM) to capture global context among tokens, ensuring that long-range dependencies are effectively modeled.
Finally, Pixel-Mamba learns a slide-level representation of the WSI, which is passed through various output heads to perform downstream tasks such as image classification, tumor staging, and survival prediction. Thanks to the linear computational complexity of the SSM and the innovative design, which includes Token Expansion and Region Fusion,  the framework supports end-to-end training with full-image WSI input, outperforming existing SOTA two-stage MIL solutions across multiple tasks.

\subsection{Whole Slide Image Serialization}
Given a whole slide image $I$ of size $H\times W \times C$, Pixel-Mamba first tokenizes $I$ into visual tokens $T = \{\tau_i, i\in [1, M]\}$,
where $H$, $W$, and $C$ represent image height, width, and channels, respectively. Here, $\tau_i$ denotes the $i^{th}$ token, and $M$ represents total number of visual tokens.
Unlike previous approaches such as Vision Mamba~\cite{zhu2024vision} or Vision Transformer~\cite{alexey2020image}, which tokenize images into fixed-size patches (e.g., $16\times 16$), Pixel-Mamba initially tokenizes each pixel individually, resulting in  $1\times 1$ pixel-wise tokens. As the tokens process through the Pixel-Mamba network, their receptive fields are progressively expanded, enabling the network to learn hierarchical representations. This is particularly important for capturing the multi-scale information inherent in pathological images.

Given the large height and width of WSIs, simple serialization strategies can cause neighboring tokens (e.g., tokens in upper and lower rows) to be far apart in the sequence, potentially disrupting the learning of inductive biases. To mitigate this issue, we employ a region-based zigzag scanning approach. As illustrated in Figure~\ref{fig:architecture} (a), the image $I$ is divided into $n$ regions of size $h \times w$, where $h$ and $w$ are the height and width of each region, respectively, and each region acts as a scanning window.
Within each scan window, a zigzag scan is applied, and a CLS token is inserted at the center of the local sequence, resulting in a sequence length of $(h \times w + 1)$ for each region.
The zigzag scan is then extended across all regions, ensuring the spatial proximity of neighboring tokens. Consequently, the total sequence length for the initial pixel-wise tokens of $I$ is $M = H \times W + n$.

\subsection{Pixel-Mamba Network}
The architecture of the hierarchical network is shown in Figure~\ref{fig:architecture} (b). 
The input to the network is the pixel-wise token series $T$ generated by the whole slide image serialization.
Each token $\tau$ initially has a size of $C=3$, corresponding to the RGB channels.
As the token receptive field expands during the forward process, the dimension $C$ is progressively increased to accommodate richer representations.
The network comprises $L=24$ Pixel-Mamba layers, each consisting of three key components: Mamba Block, Region Fusion, and Token Expansion.

Throughout the forward process, the Mamba Block models global context among tokens to capture long-range dependencies. The Region Fusion module identifies similar regions at each level and merges them to reduce redundancy, improving memory efficiency. Finally, the Token Expansion module enlarges the token receptive field, progressively increasing it from $1\times 1$ to $32 \times 32$, which is crucial for learning hierarchical representations.

\paragraph{Mamba Block}
The architecture of the Mamba Block is shown at the bottom of Figure~\ref{fig:architecture} (c). Following~\cite{zhu2024vision}, Pixel-Mamba adopts the bi-directional SSM to model the long-range dependencies of tokens at each level. 
The forward process within a Mamba Block is formulated as: 
\begin{equation}
\begin{aligned}
    T^{~l+1} &= fc(SSM_f(\psi(x))\otimes z + SSM_b(\psi(x))\otimes z) + T^{~l},\\
    x &= fc_x(norm(T^{~l})), \\
    z &= SiLU(fc_z(norm(T^{~l}))), \\
\end{aligned}
\end{equation}
where $T^{~l}$ is the token series at the $l^{th}$ layer of Pixel-Mamba, $fc$ represents the linear layer, $SSM_f$ and $SSM_b$ denote the forward and backward branches of the bi-directional SSM, respectively, and $\psi$ indicates the a Conv1D operation. The operator $\otimes$ represents element-wise multiplication, while $SiLU$ is the SiLU activation function.

\paragraph{Region Fusion}

WSIs often contain hundreds of millions of pixels, posing significant computational challenges for model training. However, much of the information in them is redundant, offering opportunities to reduce complexity. To address this, Pixel-Mamba incorporates a Region Fusion module, which iteratively fuses similar regions layer by layer, reducing redundancy and computational overhead. 
The process is illustrated at the middle of Figure \ref{fig:architecture} (c).
Given $n$ regions with $M$ tokens, the Region Fusion module partitions them into two equal sets: a gallery set and a probe set, each containing $n/2$ regions.
For each region in the probe set, the cosine similarity $S^{ij} = \frac{\text{CLS}_g^{~i} \cdot \text{CLS}_p^{~j}}{||\text{CLS}_g^{~i}||\cdot ||\text{CLS}_g^{~j}||}, i, j\in [1, n/2],$ is computed between the CLS token of this region and the CLS tokens of all regions in the gallery set,
where $\text{CLS}_g$ and $\text{CLS}_p$ represent the CLS tokens from the gallery set and probe set, respectively.

The $k$ region pairs with the highest similarity scores (Top-$k$) are then fused. Specifically, the $k$ region pairs are merged by averaging the corresponding tokens from the token series of each paired region. The number of merged regions, $k$, is defined as $\lceil {\alpha * n}/{L} \rceil$, where $0 < \alpha < 1$ is a hyper-parameter controlling the retention rate of regions for the final output, and $L$ is the total number of layers in the network. After fusion, the remaining number of regions is reduced to $n-k$.

\begin{figure}[]
  \centering
  \includegraphics[width=\columnwidth,clip=true,trim=0 90 85 0]{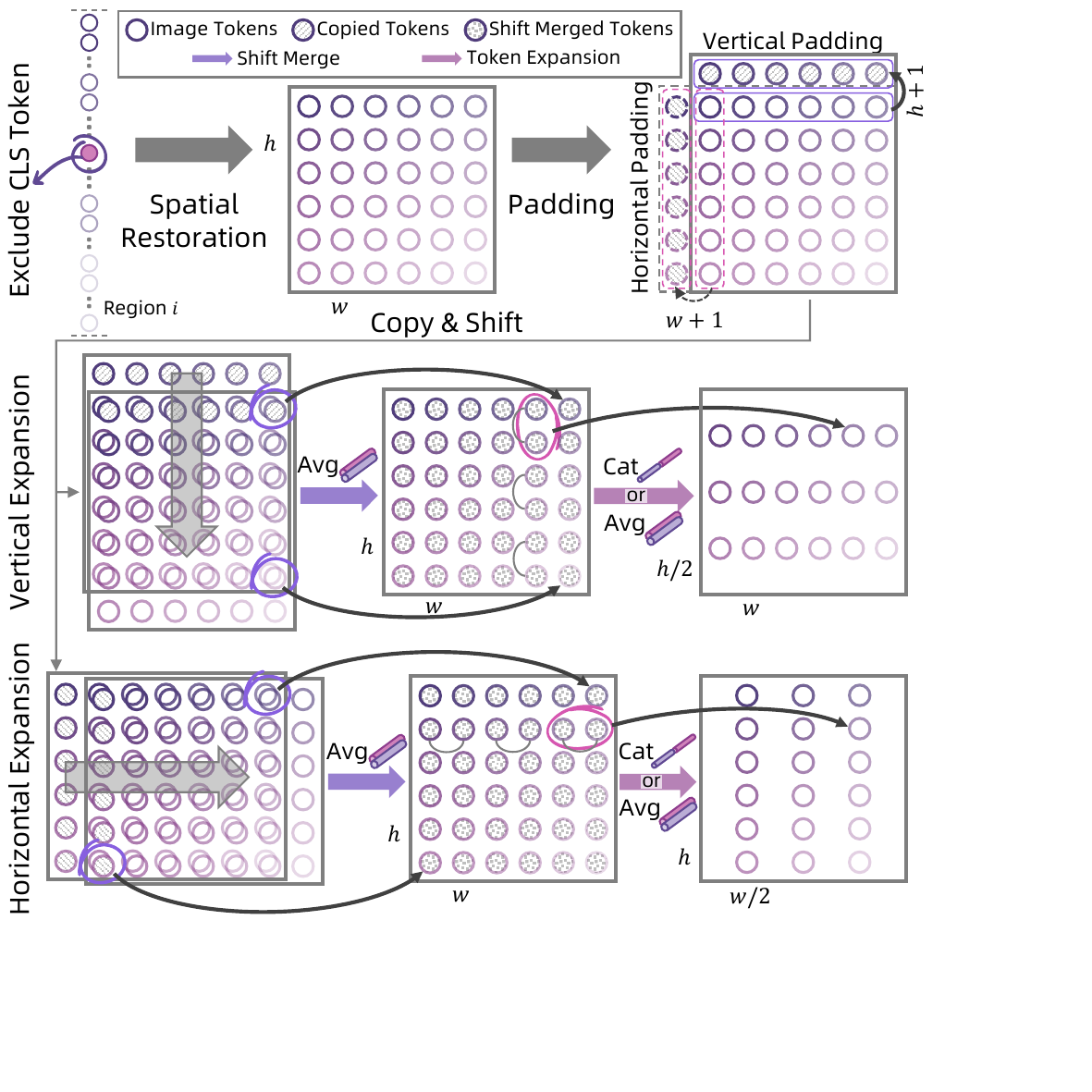}
\vspace{-0.5cm}
  \caption{The illustration of Token Expansion in a region.}
  \label{fig:token_expansion}
\end{figure}

\paragraph{Token Expansion}
Learning hierarchical token representations is critical for effective pathological image analysis. Pixel-Mamba employs a Token Expansion strategy to progressively enlarge the receptive field of tokens at key layers in the network. This enables the Mamba Block to capture multi-scale token representations at different levels.

The process of Token Expansion within a region is illustrated at the top of Figure \ref{fig:token_expansion}.
Token Expansion has two subtypes: vertical expansion and horizontal expansion, which are alternately applied across layers of the network. Given the token series of a region, the CLS token is first removed, and the remaining tokens are  reshaped into an $h \times w \times C$ feature map. . A shift merging operation is then applied, followed by token expansion. For vertical expansion, shift merging involves padding the first row, shifting the feature map down by one position, and averaging the shifted and original tokens to produce a new $h \times w \times C$ feature map with shift-merged tokens. The token expansion operation then increases the receptive field by concatenating or averaging adjacent tokens along the vertical dimension.The output size is $h/2 \times w \times 2C$ (if concatenating) or $h/2 \times w \times C$ (if averaging).

The horizontal Token Expansion follows a similar process but applies shift merging and token expansion along the $w$-dimension. The output size is $h \times w/2 \times 2C$ or $h \times w/2 \times C$, depending on the operation (concatenating or averaging).

Token Expansion progressively enlarges the receptive field of tokens as the network deepens, enabling the Mamba Block to capture multi-scale representations across levels. This process allows Pixel-Mamba to extract hierarchical representations from WSIs. The final CLS tokens from merged regions are averaged into a single token for downstream tasks. Detailed network configurations are included in the supplementary material.

\subsection{Task Heads}
Both image classification and tumor staging are formulated as classification tasks by attaching a classification head, with cross-entropy loss used for training. For survival analysis, following the formulation in \cite{zadeh2020bias}, the output CLS token from Pixel-Mamba is used to predict the hazard function $h(t)$, which is then used to compute the survival function $F(t)$. The negative log-likelihood loss $\mathcal{L}$ is defined as:
\begin{align}
\mathcal{L} = & -\sum_{(I,t,c)\in \mathcal{D}_{train}} c \cdot \text{log}(F(t|I)) \nonumber\\
&- \sum_{(I,t,c)\in \mathcal{D}_{train}}\{(1-c)\cdot \text{log}(F(t-1)|I)\\
& +(1-c)\cdot \text{log}(h(t|I))\}, \nonumber
\end{align}
where $(I,t,c)$ represents the WSI, survival time, and right-censoring status of a patient in the dataset $\mathcal{D}$.
\section{Experiments}
\label{sec:experiments}

\subsection{Implemental Details}
Implementation details for pre-training Pixel-Mamba on ImageNet and fine-tuning it on downstream datasets are provided in the supplementary materials.


\begin{table}
  \centering
  \caption{Results of image classification on ImageNet-1K~\cite{deng2009imagenet}.}
  \vspace{-0.3cm}
 \renewcommand\tabcolsep{10pt}
  \renewcommand{\arraystretch}{1.3}
  \resizebox{\columnwidth}{!}{
  \begin{tabular}{l|ccc|c}
    \toprule
    Method & Size & Params (M) & FLOPs (G) & Acc (\%) \\
    \midrule
    ResNet-18~\cite{he2016deep} & $224^2$ & 12 & 1.8 & 69.8 \\
    ResNet-50~\cite{he2016deep} & $224^2$ & 25 & 4.1 & 76.2 \\
    ResNet-101~\cite{he2016deep} & $224^2$ & 45 & 7.9 & 77.4 \\
    ResNet-152~\cite{he2016deep} & $224^2$ & 60 & 11.6 & 78.3 \\
    ResNeXt50~\cite{xie2017aggregated} & $224^2$ & 25 & 4.3& 77.6 \\
    \midrule
    ViT-B/16~\cite{alexey2020image} & $384^2$ & 86 & 55.4 & 77.9 \\
    ViT-L/16~\cite{alexey2020image} & $384^2$ & 307 & 190.7 & 76.5 \\
    DeiT-Ti~\cite{touvron2021training} & $224^2$ & 6 & 1.3 & 72.2\\
    DeiT-S~\cite{touvron2021training}  & $224^2$ & 22 & 4.6 & 79.8 \\
    \midrule
    S4ND-ViT-B~\cite{nguyen2022s4nd} & $224^2$ & 89 & - & 80.4\\
    Vim-Ti~\cite{zhu2024vision} & $224^2$ & 7.2 & 1.1 &76.1 \\
    Vim-S~\cite{zhu2024vision} & $224^2$ & 26 & 4.3 &80.5 \\

    \midrule
    \rowcolor{gray!10}
    \textbf{Pixel-Mamba-6M} & $224^2$ & 6.2 & 1.4 & 77.8 \\
    \rowcolor{gray!10}
    \textbf{Pixel-Mamba-21M}  & $224^2$ & 21 & 3.4 & 80.8\\
    \bottomrule
  \end{tabular}
  }
  \label{tab:imagenet}
\end{table}

\begin{table*}
  \centering
  \caption{The results of the tumor staging and survival analysis on pathology images. V-FM: Vision Foundation Model. VL-FM: Vision-Language Foundation Model. 
  The best results are highlighted in \textbf{bold}, and the second-best results are in \underline{underlined}.}
  \vspace{-0.3cm}
 \renewcommand\tabcolsep{3pt}
 \renewcommand{\arraystretch}{1.3}
 \resizebox{\textwidth}{!}{
 \begin{tabular}{c|c|c|ccc|ccc}
    \toprule
    \multicolumn{3}{c|}{Tasks}  &\multicolumn{3}{c|}{\textit{\textbf{Tumor Staging}} (5 Fold - Macro F1 Score)} &\multicolumn{3}{c}{\textit{\textbf{Survival Analysis}} (5 Fold - C-index)} \\
    \midrule
    Methods & Backbone & Params. & BLCA & BRCA & LUAD & BLCA & BRCA & LUAD\\
    \midrule
    \rowcolor{gray!10}
    \multicolumn{9}{c}{\textit{\textbf{Two-stage MIL}; Patch features: \textbf{ResNet-50~\cite{he2016deep}}; Pre-training: \textbf{ImageNet}}}\\
    MeanMIL & \multirow{6}{*}{ResNet-50} & 25M + 4.1K & 0.3760 $\pm$ 0.0689 & 0.1860 $\pm$ 0.0049 & 0.1800 $\pm$ 0.0089 & 0.5256 $\pm$ 0.0564 & 0.5303 $\pm$ 0.0535 & 0.5883 $\pm$ 0.0803\\
    MaxMIL & &25M + 4.1K & 0.3640 $\pm$ 0.0745 & 0.1840 $\pm$ 0.0049 & 0.1880 $\pm$ 0.0223 & 0.5250 $\pm$ 0.0548 & 0.5036 $\pm$ 0.0707 & 0.5009 $\pm$ 0.0703\\
    ABMIL~\cite{ilse2018attention} &  &25M + 0.9M & 0.4100 $\pm$ 0.0498 &0.2300 $\pm$ 0.0245& 0.2540 $\pm$ 0.0242& 0.5813 $\pm$ 0.0349 & 0.6118 $\pm$ 0.0331 & 0.6130 $\pm$ 0.0270\\
    TransMIL~\cite{shao2021transmil}&  &25M + 2.7M &0.4200 $\pm$ 0.0268 & 0.2900 $\pm$ 0.0126 & 0.3020 $\pm$ 0.0417 & 0.5610 $\pm$ 0.0223 & 0.5689 $\pm$ 0.0273 & 0.5973 $\pm$ 0.0338\\
    ILRA-MIL~\cite{xiang2023exploring} &  &25M + 3.7M& 0.4460 $\pm$ 0.0441 & 0.2400 $\pm$ 0.0261 & 0.2540 $\pm$ 0.0432 & 0.5570 $\pm$ 0.0219 & 0.5998 $\pm$ 0.0333 & 0.5725 $\pm$ 0.0519\\
    
    \midrule
    \rowcolor{gray!10}
    \multicolumn{9}{c}{\textit{\textbf{Two-stage MIL}; Patch features: \textbf{V-FM (GigaPath)~\cite{xu2024whole}}; pre-training: \textbf{171K WSIs}}}\\
    MeanMIL & \multirow{6}{*}{ViT-G} &1.1G + 6.1K& 0.4880 $\pm$ 0.0426 & 0.3160 $\pm$ 0.0287 &0.3140 $\pm$ 0.0233 & 0.6052 $\pm$ 0.0575 & 0.6255 $\pm$ 0.0391 & 0.6004 $\pm$ 0.0506\\
    MaxMIL & &1.1G + 6.1K& 0.4240 $\pm$ 0.0939 & 0.1940 $\pm$ 0.0233 & 0.2360 $\pm$ 0.0233 & 0.5798 $\pm$ 0.0290 & 0.5803 $\pm$ 0.0357 & 0.5263 $\pm$ 0.0552\\
    ABMIL~\cite{ilse2018attention} &  &1.1G + 1.2M & 0.5220 $\pm$ 0.0662& 0.3520 $\pm$ 0.0286 & 0.3820 $\pm$ 0.0331& 0.5990 $\pm$ 0.0698 & 0.6594 $\pm$ 0.0437 & 0.6083 $\pm$ 0.0461\\
    TransMIL~\cite{shao2021transmil}&  &1.1G + 2.9M & 0.4720 $\pm$ 0.0458& 0.2980 $\pm$ 0.0133& 0.3200 $\pm$ 0.0420& 0.6141 $\pm$ 0.0549 & 0.6291 $\pm$ 0.0580 & 0.5770 $\pm$ 0.0741\\
    ILRA-MIL~\cite{xiang2023exploring} &  &1.1G + 4.2M& \underline{0.5320 $\pm$ 0.0487} & 0.3640 $\pm$ 0.0294 & 0.3880 $\pm$ 0.0643 & 0.6153 $\pm$ 0.0397 & \underline{0.6528 $\pm$ 0.0430} & 0.6006 $\pm$ 0.0714\\
    \midrule
    \rowcolor{gray!10}
    \multicolumn{9}{c}{\textit{\textbf{Two-stage MIL}; Patch features: \textbf{VL-FM (CONCH)~\cite{lu2024visual}}; pre-training: \textbf{1.17 million pathology image–caption pairs}}}\\
    MeanMIL & \multirow{6}{*}{ViT-B} & 86M + 2.1K& 0.5160 $\pm$ 0.0427 & 0.3040 $\pm$ 0.0326 & 0.3260 $\pm$ 0.0459 & 0.5977 $\pm$ 0.0305 & 0.6451 $\pm$ 0.0637 & 0.6174 $\pm$ 0.0704\\
    MaxMIL &  &86M +  2.1K& 0.4660 $\pm$ 0.0408 & 0.1860 $\pm$ 0.0080 & 0.2460 $\pm$ 0.0388 & 0.5701 $\pm$ 0.0588 & 0.5483 $\pm$ 0.0742 & 0.5731 $\pm$ 0.0626\\
    ABMIL~\cite{ilse2018attention} &  &86M + 0.7M & 0.5260 $\pm$ 0.0753 & 0.3340 $\pm$ 0.0320 & 0.3840 $\pm$ 0.0700 & 0.6057 $\pm$ 0.0344 & 0.6444 $\pm$ 0.0770 & \underline{0.6399 $\pm$ 0.0578}\\
    TransMIL~\cite{shao2021transmil}&  &86M + 2.4M & 0.5180 $\pm$ 0.0312 & \underline{0.3700 $\pm$ 0.0940} & 0.3500 $\pm$ 0.0253 & \underline{0.6404 $\pm$ 0.0253} & 0.6380 $\pm$ 0.0379 & 0.5879 $\pm$ 0.0389\\
    ILRA-MIL~\cite{xiang2023exploring} &  &86M + 3.2M&0.5160 $\pm$ 0.0561 & 0.3380 $\pm$ 0.0264& \underline{0.3880 $\pm$ 0.0271} &0.6030 $\pm$ 0.0363 & 0.6532 $\pm$ 0.0409 & 0.6218 $\pm$ 0.0881 \\
    \midrule
    \rowcolor{gray!10}
    \multicolumn{9}{c}{\textit{\textbf{Two-Stage Hierarchical Representation}; pre-training: \textbf{104M pathology patches + 400K WSI regions}}}\\
     HIPT~\cite{chen2022scaling} & HIPT-ViT-6 &24M + 2.2M & 0.4660 $\pm$ 0.0185& 0.3240 $\pm$ 0.0287&0.3480 $\pm$ 0.0299 &0.5731 $\pm$ 0.0331 & 0.6139 $\pm$ 0.0446 & 0.5895 $\pm$ 0.0478 \\
    Prov-GigaPath (CLS token)~\cite{xu2024whole} & ViT-LongNet &1.2G + 6.1K& 0.4820 $\pm$ 0.0466 & 0.3060 $\pm$ 0.0388 & 0.2940 $\pm$ 0.0224 & 0.5435 $\pm$ 0.0635 & 0.5697 $\pm$ 0.0964 & 0.6044 $\pm$ 0.0294 \\
    Prov-GigaPath (feature)~\cite{xu2024whole} & ViT-LongNet &1.2G + 1.2M&0.5200 $\pm$ 0.0486 & 0.3300 $\pm$ 0.0219 & 0.3500 $\pm$ 0.0452 & 0.5954 $\pm$ 0.0311 &
0.6193 $\pm$ 0.0313 &
0.6210 $\pm$ 0.0724  \\
    \midrule
    \rowcolor{gray!10}
    \multicolumn{9}{c}{\textit{\textbf{End-to-End}; pre-training: \textbf{ImageNet + 10K WSIs}}}\\
    LongViT~\cite{wang2023image} (0.6x) & ViT-S & 22M & 0.2310 $\pm$ 0.0731 &0.3049 $\pm$ 0.0137 &0.2757 $\pm$ 0.0240 & 0.5885 $\pm$ 0.0439 & 0.6453 $\pm$ 0.0699  & 0.5890 $\pm$ 0.0236\\
    LongViT~\cite{wang2023image} (2.5x) & ViT-S & 22M & 0.4963 $\pm$ 0.0908 & 0.3068 $\pm$ 0.0276 & 0.3155 $\pm$ 0.0358 & 0.5789 $\pm$ 0.0506 & 0.6403 $\pm$ 0.0588 & 0.6085 $\pm$ 0.0140\\
    LongViT~\cite{wang2023image} (5.0x) & ViT-S & 22M & 0.4757 $\pm$ 0.0847 & 0.2979 $\pm$ 0.0265 & 0.2809 $\pm$ 0.0196  & 0.5708 $\pm$ 0.0377 & 0.6316 $\pm$ 0.0832  &0.6030 $\pm$ 0.0267   \\
    \midrule
    \rowcolor{gray!10}
    \multicolumn{9}{c}{\textit{\textbf{End-to-End}; pre-training: \textbf{ImageNet}}}\\
    \textbf{Pixel-Mamba-Stage/Surv (2.5x)} & Pixel-Mamba & 6.2M & \textbf{0.5334 $\pm$ 0.0608} & \textbf{0.3744 $\pm$ 0.0163} & \textbf{0.3917 $\pm$ 0.0125} & \textbf{0.6507 $\pm$ 0.0485} & \textbf{0.6707 $\pm$ 0.0728} & \textbf{0.6468 $\pm$ 0.0331}  \\
    \bottomrule
  \end{tabular}
  }
  \label{tab:results_main}
\end{table*}

\textbf{Comparison:}
For the comparison on natural images, we evaluate Pixel-Mamba against other networks on the ImageNet-1K classification, including classical convolutional networks (\textbf{ConvNets}), Transformer architectures (\textbf{Transformers}), and current State Space Models (\textbf{SSMs}). The representative algorithms include ResNet~\cite{he2016deep}, ViT~\cite{alexey2020image}, and Vim~\cite{zhu2024vision}.
For the comparison on pathology images, we evaluate Pixel-Mamba against other methods on tumor staging and survival analysis.
In computational pathology, most existing methods are based on a two-stage strategy of MIL, while Pixel-Mamba is an end-to-end method.
Thus, we compare the \textbf{End-to-end} framework and \textbf{Two-stage} framework on all downstream tasks.

\textbf{Two-stage:}
The two-stage framework typically consists of a feature extraction stage followed by an aggregation stage. In the first stage, many previous works employ different types of pre-trained image encoders, commonly referred to as pathological Foundation Models (FM). We select four typical FMs: 1) \textbf{ResNet-50}~\cite{he2016deep} pre-trained on ImageNet, 2) Vision Foundation Model (\textbf{V-FM: GigaPath}~\cite{xu2024whole}) pre-trained on 171K WSIs, 3) Vision Language Foundation Model (\textbf{VL-FM: CONCH}~\cite{lu2024visual}) pre-trained on 1.17 million pathological image-caption pairs, and 4) ViT with hierarchical representation (\textbf{HIPT~\cite{chen2022scaling}, Prov-GigaPath~\cite{xu2024whole}}) pre-trained on 104M pathological image patches and 400K larger WSI regions.
For the aggregation stage, we select five representative algorithms: MeanMIL, MaxMIL, ABMIL~\cite{ilse2018attention}, TransMIL~\cite{shao2021transmil}, and ILRA-MIL~\cite{xiang2023exploring}. These MIL methods are evaluated on the features extracted by the above FMs to compare the effects of different FMs.

\textbf{End-to-end:}
To the best of our knowledge, only LongViT\cite{wang2023image} has achieved end-to-end training on WSIs using a well-designed dilated attention network (LongNet)\cite{ding2023longnet}. However, it requires 64 A100 GPUs for training, which is impractical for most researchers. We also compare LongViT with Pixel-Mamba across all downstream tasks.

\subsection{Image Classification on Natural Images}
\textbf{Settings:}
We pre-train and evaluate Pixel-Mamba on natural images of ImageNet-1K~\cite{deng2009imagenet}, which includes more than 1.2 million images for training and 50k images for validation, covering 1000 different categories.
Following previous works~\cite{touvron2021training, zhu2024vision}, we use top-1 accuracy for evaluation.

\paragraph{Results:}
The results of image classification are summarized in Table \ref{tab:imagenet}. We pre-trained Pixel-Mamba in two configurations: Pixel-Mamba-6M and Pixel-Mamba-21M. Overall, compared to ConvNets, Transformers, and SSMs, Pixel-Mamba achieves the highest accuracies of 77.8\% and 80.8\% for the respective configurations. In particular, with an architecture similar to SSM, Pixel-Mamba-6M outperforms Vim-Ti by 1.7\% in accuracy while using fewer model parameters. These results demonstrate that Pixel-Mamba has a stronger capacity than other methods and lays a better pre-training foundation for downstream tasks.

\subsection{Tumor Staging on Pathology Images}
\noindent \textbf{Settings:}
For tumor staging, we use Pixel-Mamba as the backbone network and add a linear layer as the staging head, which projects features into probabilities for four stages. This staging model is named Pixel-Mamba-Stage. Three datasets from The Cancer Genome Atlas (TCGA) are employed: Bladder Urothelial Carcinoma (BLCA), Breast Invasive Carcinoma (BRCA), and Lung Adenocarcinoma (LUAD). Each patient is classified into one of four tumor stages. Following previous work~\cite{li2024dynamic}, due to the imbalanced number of patients in each stage, we use the macro-F1 score with 5-fold cross-validation for evaluation.

\paragraph{Results:}
The results of tumor staging are summarized in Table \ref{tab:results_main}. Compared to different FMs such as ResNet-50, GigaViT, and CONCH, a better pre-trained FM significantly improves the performance of two-stage MIL approaches. Both vision FM (V-FM, GigaPath) and vision-language FM (VL-FM, CONCH) demonstrate comparable results. Overall, Pixel-Mamba-Stage outperforms two-stage MIL methods, two-stage hierarchical representation methods, and LongViT with macro F1 scores of 0.5334, 0.3744, and 0.3917 on the BLCA, BRCA, and LUAD datasets, respectively. Besides, all two-stage methods are pre-trained on large-scale pathological images or image-caption pairs, whereas LongViT is pre-trained on 10k WSIs. In contrast, Pixel-Mamba-Stage is only pre-trained on natural images, showcasing the efficacy of Pixel-Mamba.

\subsection{Survival Analysis on Pathology Images}
\paragraph{Settings:}
For survival analysis, we use Pixel-Mamba as the backbone network and add a linear layer as the survival head, which predicts the risk of patients, named Pixel-Mamba-Surv. The datasets used for survival analysis are BLCA, BRCA, and LUAD.
Following previous work~\cite{chen2022scaling,xu2024whole}, we use the Concordance index (C-index) with 5-fold cross-validation for evaluation.

\paragraph{Results:}
The results of survival analysis are summarized in Table \ref{tab:results_main}. Similar to tumor staging, using pre-trained GigaPath and CONCH as feature extractors, two-stage MIL approaches achieve better results in C-index compared to using ResNet-50. Pixel-Mamba-Surv, pre-trained on only natural images, outperforms all two-stage MIL methods, two-stage hierarchical methods, and the end-to-end LongViT on the BLCA, BRCA, and LUAD datasets, with C-index values of 0.6507, 0.6707, and 0.6468, respectively. These results demonstrate that learning slide representations in an end-to-end manner can surpass existing two-stage methods, even without pre-training on pathological images.

\begin{figure}[t]
  \centering
  \includegraphics[width=\columnwidth]{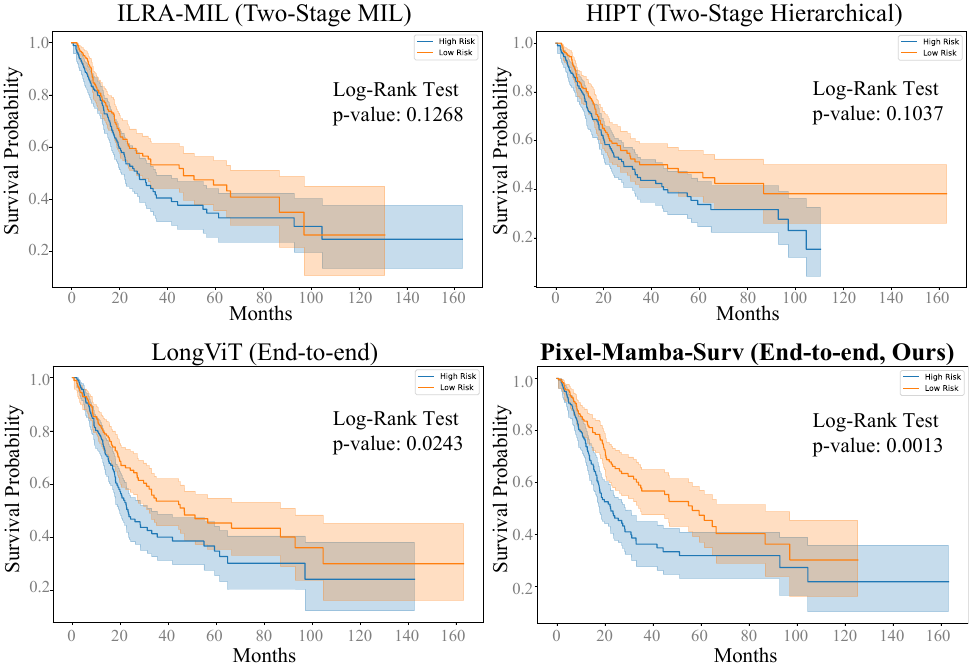}
\vspace{-0.5cm}
  \caption{The comparison of Kaplan-Meier analysis and Log-Rank test (p-value, lower is best and p-value $\leq$ 0.01 indicates the statistical significance between two groups) on the BLCA dataset.}
  \label{fig:km}
\end{figure}

In addition, we analyze the performance of four different methods (ILRA-MIL, HIPT, LongViT, and Pixel-Mamba-Surv) using Kaplan-Meier analysis and the Log-Rank test to measure the statistical significance between low-risk and high-risk groups. The Kaplan-Meier curves after stratifying patients into low and high-risk categories based on their median predicted risk scores, are shown in Figure \ref{fig:km}. Pixel-Mamba-Surv demonstrates better separation between the low-risk and high-risk groups compared to the other methods, and its lower p-value (0.0013) indicates statistically significant predictions.




\subsection{Ablation Study}

\paragraph{Two-stage \textit{vs} End-to-end in WSIs:}
As shown in Table \ref{tab:results_main}, although two-stage MIL approaches with pre-trained FMs achieve better results than end-to-end LongViT, end-to-end Pixel-Mamba-Stage/Surv without pre-training on pathological images outperforms both two-stage MIL approaches and LongViT.
To evaluate the end-to-end Pixel-Mamba, we transferred LongViT and Vision Mamba~\cite{zhu2024vision} into the survival prediction on the BLCA dataset. As shown in Table \ref{tab:end2end}, Vision-Mamba-Surv outperforms LongViT-Surv by a relative 4.1\% in C-index with fewer parameters. This demonstrates that the SSM-based long-sequence modeling scheme is more advantageous than the optimized dilated attention mechanism in LongViT when processing ultra-long tokens in WSIs. Furthermore, Pixel-Mamba-Surv with hierarchical slide representation outperforms Vision-Mamba-Surv by a relative 8.9\% in C-index. This indicates that Pixel-Mamba, with its hierarchical representation and end-to-end optimization, is better at handling pathological images.

\begin{table}
  \centering
  \caption{Ablation study of end-to-end survival analysis on BLCA dataset. The used WSIs are at 2.5x magnification.}
  \vspace{-0.3cm}
 \renewcommand\tabcolsep{4pt}
  \renewcommand{\arraystretch}{1.3}
  \resizebox{\columnwidth}{!}{
  \begin{tabular}{c|cc|c}
    \toprule
    Method & Backbone & Params(M)  & C-index \\
    \midrule
    LongViT-Surv & ViT-S& 22 & 0.5743 $\pm$ 0.0236\\
    Vision-Mamba-Surv & Vim-Ti & 7.2 & 0.5977 $\pm$ 0.0129\\
    {Pixel-Mamba-Surv} & Pixel-Mamba& 6.2 & {0.6507 $\pm$ 0.0485}\\
    \bottomrule
  \end{tabular}
  }
  \label{tab:end2end}
\end{table}

\paragraph{Hierarchical Tokenization:}
The token scales of existing two-stage methods or end-to-end LongViT are fixed (e.g., 16, 32), whereas hierarchical representation is crucial for WSI analysis. In Table \ref{tab:token}, we present experiments using Vision-Mamba-Surv on two independent token scales: 16 and 32, achieving C-index values of 0.5977 and 0.5859, respectively. When combining the two tokens, Vision-Mamba-Surv achieves a higher C-index of 0.6116, indicating that multi-scale token representation is effective for WSI analysis. Moreover, utilizing hierarchical token representation with scales ranging from 1 to 32, Pixel-Mamba-Surv achieves a C-index of 0.6507. These results demonstrate that the proposed hierarchical representation is effective for pathological image analysis.

\begin{table}
  \centering
  \caption{Ablation study of hierarchical tokenization on BLCA dataset. Hier-(1-32) indicates that the tokens of Pixel-Mamba are hierarchical, with token scales ranging from 1 to 32. }
  \vspace{-0.3cm}
 \renewcommand\tabcolsep{4pt}
  \renewcommand{\arraystretch}{1.3}
  \resizebox{\columnwidth}{!}{
  \begin{tabular}{l|cc|c}
    \toprule
    Method & Backbone & Token Scale & C-index \\
    \midrule
    Vision-Mamba-Surv & Vim-Ti & 16 & 0.5977 $\pm$ 0.0129\\
    Vision-Mamba-Surv & Vim-Ti & 32 & 0.5859 $\pm$ 0.0470\\
    Vision-Mamba-Surv & Vim-Ti & 16 + 32 & 0.6116 $\pm$ 0.0313\\
    \midrule
    Pixel-Mamba-Surv & Pixel-Mamba & Hier-(1-32) & 0.6507 $\pm$ 0.0485\\
    \bottomrule
  \end{tabular}
  }
  \label{tab:token}
\end{table}

\paragraph{Scan Window of Pixel-Mamba:}
Pathological images often exceed 10,000 pixels in length and width, resulting in the initial number of Pixel-Mamba tokens surpassing 100 million. Directly applying the sequence scanning method of Vision Mamba would result in excessively long distances between spatially adjacent tokens. Inspired by Local Mamba~\cite{huang2024localmamba}, Pixel-Mamba introduces the scan window for processing pathological tokens, reducing the sequence distance between spatially adjacent tokens and enhancing the learning of local inductive biases.
As shown in Table \ref{tab:window}, Pixel-Mamba-Surv achieves the best results in the scan window size of 224 $\times$ 224 since it is pre-trained in this size.

\begin{table}
  \centering
  \caption{Ablation study of scan window on BLCA dataset.}
  \vspace{-0.3cm}
 \renewcommand\tabcolsep{4pt}
  \renewcommand{\arraystretch}{1.3}
  \resizebox{\columnwidth}{!}{
  \begin{tabular}{l|c|c|c}
    \toprule
    Method & Magnification & Scan window & C-index \\
    \midrule
    Pixel-Mamba-Surv & 2.5x & 32 & 0.6083 $\pm$ 0.0562\\
    Pixel-Mamba-Surv & 2.5x & 128 & 0.6190 $\pm$ 0.0374\\
    Pixel-Mamba-Surv & 2.5x & 224 & 0.6507 $\pm$ 0.0485\\
    Pixel-Mamba-Surv & 2.5x & 320 & 0.6242 $\pm$ 0.0386\\
    \bottomrule
  \end{tabular}
  }
  \label{tab:window}
\end{table}

\paragraph{Magnification of WSIs:}
Existing two-stage methods~\cite{xu2024whole,chen2024uni} typically divide WSIs at 20x magnification into tens of thousands of tiles, capturing more fine-grained cellular details. Similar to LongViT~\cite{wang2023image}, the end-to-end Pixel-Mamba can also adapt to images of various sizes. As shown in Table \ref{tab:wsi_scale}, with increasing input size, Pixel-Mamba-Surv achieves better results, realizing a C-index of 0.6507 on the BLCA dataset. These results demonstrate its scalability with the increasing number of tokens. Limited by GPU memory, the current Pixel-Mamba supports end-to-end training for WSIs at 2.5x magnification. Achieving end-to-end training on larger sizes will be our future work.

\begin{table}
  \centering
  \caption{Ablation study of image scale on BLCA dataset.}
  \vspace{-0.3cm}
 \renewcommand\tabcolsep{4pt}
  \renewcommand{\arraystretch}{1.3}
  \resizebox{\columnwidth}{!}{
  \begin{tabular}{l|cc|c}
    \toprule
    Method & Backbone & Magnification & C-index \\
    \midrule
    Pixel-Mamba-Surv & Pixel-Mamba & 0.6x & 0.5855 $\pm$ 0.0286\\
    Pixel-Mamba-Surv & Pixel-Mamba & 1.0x & 0.6307 $\pm$ 0.0500\\
    Pixel-Mamba-Surv & Pixel-Mamba & 2.5x & 0.6507 $\pm$ 0.0485\\
    
    \bottomrule
  \end{tabular}
  }
  \label{tab:wsi_scale}
\end{table}

\textbf{Discussion:}
Pixel-Mamba showcases the significant potential of hierarchical representation and end-to-end optimization in WSI analysis.
Pixel-Mamba's lightweight design, even without pre-training on pathology data, surpasses existing foundation models across multiple downstream tasks. 
This study emphasizes the importance of slide representation and end-to-end optimization. 
Future directions for Pixel-Mamba include: 1) gathering a sufficient number of WSIs for the pre-training of slide representation, and 2) optimizing the model structure to enable end-to-end training on 10x/20x magnified images to capture fine-grained details.

\section{Conclusion}
\label{sec:conclusion}
In this paper, we propose Pixel-Mamba, a novel end-to-end and practical framework with state space model (Mamba), for learning hierarchical slide representations of whole slide image from pixels to gigapixels. Current multiple-instance learning methods rely heavily on the pre-training of foundation models with extensive histopathology data, while Pixel-Mamba achieves the results that surpass existing state-of-the-art multiple-instance learning methods in many downstream tasks, even without the pre-training on histopathology images. We believe that Pixel-Mamba offers a new and practical end-to-end optimization solution and can be a simple baseline for end-to-end histopathology image analysis, marking an important milestone in the advancement of histopathology image analysis by end-to-end whole slide representation learning.

\appendix
\section{Appendix}
\subsection{Network Configurations of Pixel-Mamba}
\label{sec:network}
The detailed network configurations of Pixel-Mamba-6M and Pixel-Mamba-21M are provided in Table \ref{tab:pixel_mamba_6m} and Table \ref{tab:pixel_mamba_21m}, respectively.

Take Pixel-Mamba-6M for example, it consists of 24 Pixel-Mamba layers,  each composed of three key components: Mamba Block, Region Fusion (RF), and Token Expansion (TE).
Throughout the forward process, the Mamba Block models global context among tokens to capture long-range dependencies. The Region Fusion module identifies similar regions at each level and merges them to reduce redundancy, improving memory efficiency. Finally, the Token Expansion module enlarges the token receptive field, progressively increasing it from 1 × 1 to 32 × 32, which is crucial for learning hierarchical representations.
With the expansion of the token receptive field, the feature channels of tokens are increasing from 3 to 384.
Pixel-Mamba includes horizontal TE and vertical TE. In each one, the token expansion operation then increases the receptive field by concatenating or averaging adjacent tokens along the horizontal or vertical dimensions.

Pixel-Mamba-21M also consists of 24 Pixel-Mamba layers. The difference is that its token feature channel increases to 768 throughout the forward process.

\begin{table}
  \centering
  \caption{The detailed network configurations of Pixel-Mamba-6M. In the column of Pixel-Mamba Layers, Mamba represents the Mamba Block, RF represents the Region Fusion, and TE represents Token Expansion.
  The Token column indicates the receptive field of tokens in this layer.
  }
 \renewcommand\tabcolsep{4pt}
  \renewcommand{\arraystretch}{1.3}
  \resizebox{\columnwidth}{!}{
  \begin{tabular}{c|c|c|c}
    \toprule
    \multicolumn{4}{c}{\textbf{Pixel-Mamba-6M}}\\
    \midrule
    Layer id & Token & Channels & Pixel-Mamba Layers \\
    \midrule
    1 & $1\times 1$ & 3 & Mamba + RF + Horizontal TE-Cat \\
    2 & $1\times 2$ & 6 & Mamba + RF + Vertical TE-Cat \\
    3 & $2\times 2$& 12 & Mamba + RF + Horizontal TE-Cat \\
    4 & $2\times 4$& 24 & Mamba + RF + Vertical TE-Cat\\
    5 & $4\times 4$& 48 & Mamba + RF\\
    6 & $4\times 4$& 48 & Mamba + RF + Horizontal TE-Cat\\
    7 & $4\times 8$& 96 & Mamba + RF\\
    8 & $4\times 8$& 96 & Mamba + RF + Vertical TE-Cat\\
    9 & $8\times 8$& 192 & Mamba + RF\\
    10 & $8\times 8$&192 & Mamba + RF\\
    11 & $8\times 8$&192 & Mamba + RF + Horizontal TE-Avg\\
    12 & $8\times 16$&192 & Mamba + RF \\
    13 & $8\times 16$&192 & Mamba + RF\\
    14 & $8\times 16$&192 & Mamba + RF + Vertical TE-Avg\\
    15 & $16\times 16$&192 & Mamba + RF\\
    16 & $16\times 16$&192 & Mamba + RF\\
    17 & $16\times 16$&192 & Mamba + RF\\
    18 & $16\times 16$&192 & Mamba + RF\\
    19 & $16\times 16$&192 & Mamba + RF\\
    20 & $16\times 16$&192 & Mamba + RF + Horizontal TE-Avg\\
    21 & $16\times 32$&192 & Mamba + RF \\
    22 & $16\times 32$&192 & Mamba + RF + Vertical TE-Cat\\
    23 & $32\times 32$&384 & Mamba + RF\\
    24 & $32\times 32$&384 & Mamba + RF\\
    \bottomrule
  \end{tabular}
  }
  \label{tab:pixel_mamba_6m}
\end{table}

\begin{table}
  \centering
  \caption{The detailed network configurations of Pixel-Mamba-21M. In the column of Pixel-Mamba Layers, Mamba represents the Mamba Block, RF represents the Region Fusion, and TE represents Token Expansion.
  The Token column indicates the receptive field of tokens in this layer.
  }
 \renewcommand\tabcolsep{4pt}
  \renewcommand{\arraystretch}{1.3}
  \resizebox{\columnwidth}{!}{
  \begin{tabular}{c|c|c|c}
    \toprule
    \multicolumn{4}{c}{\textbf{Pixel-Mamba-21M}}\\
    \midrule
    Layer id & Token & Channels & Pixel-Mamba Layers \\
    \midrule
    1 & $1\times 1$ & 3 & Mamba + RF + Horizontal TE-Cat \\
    2 & $1\times 2$ & 6 & Mamba + RF + Vertical TE-Cat \\
    3 & $2\times 2$& 12 & Mamba + RF + Horizontal TE-Cat \\
    4 & $2\times 4$& 24 & Mamba + RF + Vertical TE-Cat\\
    5 & $4\times 4$& 48 & Mamba + RF\\
    6 & $4\times 4$& 48 & Mamba + RF + Horizontal TE-Cat\\
    7 & $4\times 8$& 96 & Mamba + RF\\
    8 & $4\times 8$& 96 & Mamba + RF + Vertical TE-Cat\\
    9 & $8\times 8$& 192 & Mamba + RF\\
    10 & $8\times 8$&192 & Mamba + RF\\
    11 & $8\times 8$&192 & Mamba + RF + Horizontal TE-Cat\\
    12 & $8\times 16$&384 & Mamba + RF \\
    13 & $8\times 16$&384 & Mamba + RF\\
    14 & $8\times 16$&384 & Mamba + RF + Vertical TE-Avg\\
    15 & $16\times 16$&384 & Mamba + RF\\
    16 & $16\times 16$&384 & Mamba + RF\\
    17 & $16\times 16$&384 & Mamba + RF\\
    18 & $16\times 16$&384 & Mamba + RF\\
    19 & $16\times 16$&384 & Mamba + RF\\
    20 & $16\times 16$&384 & Mamba + RF + Horizontal TE-Avg\\
    21 & $16\times 32$&384 & Mamba + RF \\
    22 & $16\times 32$&384 & Mamba + RF + Vertical TE-Cat\\
    23 & $32\times 32$&768 & Mamba + RF \\
    24 & $32\times 32$&768 & Mamba + RF\\
    \bottomrule
  \end{tabular}
  }
  \label{tab:pixel_mamba_21m}
\end{table}

\begin{table}
  \centering
  \caption{The ablation study of $\alpha$ in Region Fusion.
  }
 \renewcommand\tabcolsep{12pt}
  \renewcommand{\arraystretch}{1.3}
  \resizebox{\columnwidth}{!}{
  \begin{tabular}{c|c|c|c|c}
    \toprule
    $\alpha$ & 0.4 & 0.6 & 0.8 & 1.0 \\
    \midrule
    C-index & 0.6176 & 0.6260 & 0.6507 & 0.6104 \\
    \bottomrule
  \end{tabular}
  }
  \label{tab:ab_a}
\end{table}

\subsection{Implemental Details}
\label{sec:implemental}
\paragraph{Pre-training:}
Pixel-Mamba is pre-trained on the ImageNet~\cite{deng2009imagenet} to ensure model convergence, utilizing over 1.28 million natural images for a classification task encompassing 1,000 categories.
Following \cite{zhu2024vision}, Pixel-Mamba is trained on images of size 224×224. The optimizer is AdamW with a momentum of 0.9, batch size of 1024, and weight decay of 0.05. Pixel-Mamba is trained for 300 epochs using a cosine learning rate schedule, starting with an initial learning rate of 0.001. Data augmentation techniques include random cropping, random horizontal flipping, label-smoothing regularization, mixup, and random erasing. 8 Nvidia A100 GPUs are used for pre-training, costing approximately 2 to 3 days.

\paragraph{Fine-tuning on Downstream Tasks:}
Pixel-Mamba serves as the backbone network for all downstream tasks involving pathology images, including tumor staging and survival analysis. During fine-tuning, a task-specific head is added to the backbone network, and the entire network is fine-tuned on WSIs at a 2.5$\times$ magnification over 100 epochs. The AdamW optimizer is employed alongside a cosine learning rate schedule, starting with an initial learning rate of 0.0004. 8 Nvidia A100 GPUs are used. In each iteration, one GPU is assigned to process one WSI. Gradient accumulation is performed every 8 iterations, resulting in an effective batch size of 64.

\subsection{The ablation of Region Fusion}
\label{sec:ab_region_fusion}
In the Region Fusion module of Pixel-Mamba, the number of merged regions, $k$, is defined as $\lceil {\alpha * n}/{L} \rceil$, where $0 < \alpha < 1$ is a hyper-parameter controlling the retention rate of regions for the final output, and $L$ is the total number of layers in the network.

We conduct the ablation study of hyper-parameter $\alpha$ and the results are reported in Table \ref{tab:ab_a}.
Pixel-Mamba-Surv achieves the best C-index of 0.6507 with the $\alpha = 0.8$ on the BLCA dataset. Thus, we suggest $\alpha = 0.8$, and all results of experiments in the main text of the manuscript are obtained with $\alpha = 0.8$.

\clearpage
{
    \small
    \bibliographystyle{ieeenat_fullname}
    \bibliography{main}
}



\end{document}